%% file: Box_EN_ver_SimiFinal.tex
\documentclass[journal, twocolumn, 10pt, table]{IEEEtran}

\usepackage{graphicx}
\usepackage{amsmath, amsthm, amssymb, bm}
\usepackage{amsthm}
\usepackage{algpseudocode}
\usepackage{epstopdf}
\usepackage{color}
\usepackage{subfigure}
\usepackage{epsfig}
\usepackage{flushend}
\usepackage{cleveref}
\usepackage{color}

\newtheorem{theorem}{Theorem}

\usepackage[linesnumbered,ruled]{algorithm2e}

\newcommand{\skipval}{0.901mm} 
\usepackage{amsmath}
\usepackage{epstopdf}
\usepackage{amssymb}
\usepackage{systeme}
\usepackage{color}
\usepackage{pgfplots}
\usepackage{bbm}

\include{macros}
\ifCLASSINFOpdf
\else
\fi
\begin{document}
%

\title{Precise Performance Analysis of the Box-Elastic Net under Matrix Uncertainties}
%
%
%

%

\author{Ayed~M.~Alrashdi, Ismail~Ben~Atitallah, and Tareq~Y.~Al-Naffouri

\vspace{-12pt}

\thanks{A. M. Alrashdi is with the Computer, Electrical, and Mathematical Sciences
and Engineering Division, King Abdullah University of Science and
Technology, Thuwal 23955, Saudi Arabia, and also with the Department of
Electrical Engineering, University of Hail, Hail 55476, Saudi Arabia (e-mail:
ayed.alrashdi@kaust.edu.sa).

Ismail Ben Atitallah is with the Department of Electrical Engineering, Harvard University, USA. (e-mail: benatitallah@g.harvard.edu).

T. Y. Al-Naffouri is with the Computer, Electrical, and Mathematical Sciences and Engineering (CEMSE) Division, King Abdullah University of Science and Technology (KAUST), Thuwal 23955, Saudi Arabia. (e-mail: tareq.alnaffouri@kaust.edu.sa).
}
}

\maketitle

\begin{abstract}
In this letter, we consider the problem of recovering an unknown sparse signal from noisy linear measurements, using an enhanced version of the popular Elastic-Net (EN) method. We modify the EN by adding a box-constraint, and we call it the Box-Elastic Net (Box-EN). We assume independent identically distributed (iid) real Gaussian measurement matrix with additive Gaussian noise.
In many practical situations, the measurement matrix is not perfectly known, and so we only have a noisy estimate of it. 
In this work, we precisely characterize the mean squared error and the probability of support recovery of the Box-Elastic Net in the high-dimensional asymptotic regime. 
Numerical simulations validate the theoretical predictions derived in the paper and also show that the boxed variant outperforms the standard EN.
\end{abstract}

\IEEEpeerreviewmaketitle
\section{Introduction}
\label{sec:intro}
The Elastic-Net (EN) \cite{zou2005regularization} is a powerful method to recover an unknown signal $\xv_0$ from noisy linear measurements
$\yv = \Hm \xv_{0} + \zv$ 
by solving the following optimization problem: 
\begin{equation}
\label{EN_1}   
\hat{\xv}=  \text{arg} \ \underset{{\xv \in \mathbb{R}^{n}}}{\operatorname{\min}} \ \|  \yv - \Hm \xv \|_2^{2} + \lambda_1 \| \xv \|_1+ \lambda_2 \| \xv \|_2^2,
\end{equation}
where $ \Hm$ is the measurement matrix, $\zv$ is the noise vector, and $\lambda_1, \lambda_2 >0$ are the regularization parameters.
The EN reduces to the ridge regression for $\lambda_1 \to 0$, and to the LASSO \cite{tibshirani1996regression} as $\lambda_2 \to 0$. It combines the good features of both of these approaches and overcomes many limitations of the LASSO \cite{zou2005regularization}.
This combination allows for learning a sparse model where few of the entries are non-zero like the LASSO, while still maintaining the regularization properties of the ridge regression.
The EN has been used in many applications \cite{hastie2015statistical}, \cite{wang2006doubly}, \cite{shen2014doubly}, \cite{montanari2017universality}, \cite{khalajmehrabadi2017joint}, \cite{you2016oracle}.
In this paper, we refer to (\ref{EN_1}) as the \textit{standard} EN, but we focus on a modified version that we call the \textit{Box-EN} which solves the following optimization instead
\begin{equation}\label{Box-EN_1} 
\hat{\xv}=  \text{arg} \ \underset{{\xv \in \mathcal{B}}}{\operatorname{\min}} \ \|  \yv - \Hm \xv \|_2^{2} + \lambda_1 \| \xv \|_1+ \lambda_2 \| \xv \|_2^2,
\end{equation}
                   $ \text{where}, \mathcal{B} = [l, u]^n, \text{and} \quad l \leq 0, u \geq 0 \in \mathbb{R}.$

The box constraint, which can be seen as an $\ell_{\infty}$-norm constraint, is used to promote a boundedness constraint. For instance, if the EN aims to recover a digital bounded signal through relaxation\footnote{without the relaxation, having a discrete feasible set lead to a computationally prohibitive algorithm.}, then it is potentially useful to add the box constraint, thereby yielding the Box-EN.
When $l=-u$, the Box-EN is equivalent to Regularized Least Squares, where we use three regularization norms: $\ell_1$, $\ell_2$ and $\ell_{\infty}$. This is used to solve simultaneously structured signals \cite{oymak2015simultaneously}, for instance signals that are known to be both sparse and bounded, in a high-dimensional setting.
Comparing (\ref{Box-EN_1}) to (\ref{EN_1}), the only small variation is the "box-constraint'' that has been added to (\ref{Box-EN_1}). However, as we will see later, this small variation assures significant improvement in the performance in scenarios where the entries of $\xv_0$ are bounded or approximately so. Examples of such signals are found in many applications such as wireless communication systems \cite{jeganathan2009space}, image processing \cite{ting2009sparse}, etc.
The Box-EN is not as popular as the standard EN, but we can find some references where different types of constraints were imposed on the algorithm \cite{rodola2013elastic}, \cite{li2015constrained}, \cite{liu2016robust}.

In the last few years, various forms of precise analysis of the asymptotic estimation error of non-smooth regularized convex optimization problems (such as LASSO and related inverse problems) have been proved under the assumption of noisy iid Gaussian measurements. They mainly follow one of two parallel approaches.
The first one is the Approximate Message Passing (AMP) framework that has been used in \cite{donoho2009message, bayati2011dynamics, bayati2012lasso} to derive precise asymptotic analysis of the LASSO performance under the assumptions of iid Gaussian sensing matrix. The second approach is based on a recently developed framework that uses the Convex Gaussian Min-max Theorem (CGMT) \cite{thrampoulidis2018precise}. It has been used in a series of works to also precisely evaluate the estimation performance of these problems under the same Gaussianity assumptions \cite{thrampoulidis2018symbol,thrampoulidis2015regularized, atitallah2017box}.

However, these results assume that the measurement matrix $\Hm$ is perfectly known. In many practical applications it is reasonable to expect uncertainty in the measurement matrix due to, e.g., imperfections in the signal acquisition hardware, model mismatch, or estimation errors \cite{rosenbaum2010sparse}.
In this paper, we consider the \textit{additive} uncertainty model: $\Am = \sqrt{1-\epsilon^2} \Hm + \epsilon \Omegam,$ where $\Hm$ is known, $\Omegam$ is an unknown error matrix, and $\epsilon^2\in [0,1]$ is the variance of the error. Such model is commonly used in communication theory and known as imperfect Channel State Information (CSI) \cite{zenaidi2016performance}, \cite{alrashdi2017precise},\cite{alrashdi2018optimum}.

In this work, we derive novel precise asymptotic (in the problem dimensions $m$ and $n$) characterizations of the \textit{mean squared error} and the \textit{support recovery probability} of the Box-EN under the presence of uncertainties in the measurement matrix that has iid Gaussian entries using the CGMT framework. 
Although our analysis is asymptotic in nature, numerical simulations show that our predictions are valid even for a few dozens of the problem dimensions. They also demonstrate that the Box-EN outperforms the standard one. These predictions can be used to optimally tune the involved parameters of the algorithm.
To the best of our knowledge, the precise error analysis of the Box-EN under uncertainties has not been explicitly derived before. 
Finally, we note that following the same steps as in this paper, our results can be used to characterize the performance of the standard EN as well.
\section{Problem Setup}
\subsection{System Model}
\label{w_A}
We consider the problem of recovering an unknown signal $\xv_{0} \in \mathbb{R}^{n}$ from a noisy linear measurement vector $\yv = \Hm \xv_0 +\zv$.
The unknown signal vector $\xv_{0}$ is assumed to be $k$-sparse, i.e., only $k$ of its entries are sampled iid from a distribution $ p_{X_0}$ which has zero mean and unit variance ($\mathbb{E}[X_0^2] = 1$), and the remaining entries are zeros. 
For the measurement matrix, we consider the following additive uncertainty model:
$\Am = \gamma \Hm + \epsilon \Omegam,$ 
where $\Hm$, and $\Omegam \in \mathbb{R}^{m \times n}$ have iid entries $\mathcal{N}(0,1/n)$,  
 and $\epsilon^2 \in [0,1]$ is the variance of the error such that $\gamma^2 +\epsilon^2 = 1$.
The noise vector $\zv \in \mathbb{R}^{m}$ has entries iid $\mathcal{N}(0,\sigma_{\zv}^2)$.
The analysis is performed when the system dimensions ($m$, $n$ and $k$) grow simultaneously large at fixed ratios:
$\frac{m}{n} \longrightarrow \delta \in (0,\infty)$, and $\frac{k}{n} \longrightarrow \kappa \in (0,1).$ Under these settings, the signal-to-noise ratio (SNR) becomes SNR := $\frac{\kappa}{\sigma_{\zv}^2}$.
\subsection{Performance Metrics}
We consider the following two performance metrics of the Box-Elastic Net:\\
\textbf{Mean squared error (MSE)}: A natural and heavily used measure of performance is the reconstruction \textit{mean squared error}, which measures the deviation of $\hat{\xv}$ from the true signal $\xv_0$. Formally, the MSE is defined as $\text{MSE} := \frac{1}{n}\| \hat{\xv} - \xv_0 \|_2^2$.\\
\textbf{Support Recovery}: In the problem of sparse recovery, a natural measure of performance that is used in many applications 
 is support recovery, which is defined as identifying whether an entry of $\xv_0$ is on the support (i.e., non-zero), or it is off the support (i.e., zero). The decision is based on the Box-EN solution $\hat{\xv}$: we say the $i^{th}$ entry of $\hat{\xv}$ is on the support if $| \hat{\xv}_{i}| \geq \xi$, where $\xi > 0$ is a user-defined hard threshold on the entries of $\hat{\xv}.$ Formally, let
\begin{subequations}\label{supp}
\begin{align}
\Phi_{\xi,\text{on}}(\hat{\xv}) = \frac{1}{k} \sum_{i \in S(\xv_0)} \mathbbm{1}_{\{| \hat{\xv}_{i}| \geq \xi \}},\\
\Phi_{\xi,\text{off}}(\hat{\xv}) = \frac{1}{n-k} \sum_{i \notin S(\xv_0)} \mathbbm{1}_{\{| \hat{\xv}_{i}| \leq \xi \}},
\end{align}
\end{subequations}
where $\mathbbm{1}_{\{\mathcal{A} \}}$ is the indicator function of a set $\mathcal{A}$, and $S(\xv_0)$ is the support of $\xv_0$, i.e., the set of the non-zero entries of $\xv_0$. In Theorem \ref{EN_on/off}, we precisely predict the \textit{per-entry} rate of successful on-support and off-support recovery.

\section{Approach and Sketch of the Proof}
\label{Proof}
In this section, we provide the asymptotic analysis of the MSE of the proposed Box-EN optimization. Our approach is based on the recently developed Convex Gaussian Min-max Theorem (CGMT) which is summarized in the next subsection.
\vspace{-0.5cm}
\subsection{Convex Gaussian Min-max Theorem (CGMT)}
We first need to state the key ingredient of the analysis which is the Convex Gaussian Min-max Theorem CGMT. Here, we recall the statement of the theorem, and we refer the reader to \cite{thrampoulidis2018precise} for the complete technical requirements.
Consider the following two min-max problems, which we refer to as the Primary Optimization (PO) and Auxiliary Optimization (AO):
\begin{subequations}
\begin{align}\label{P,AO}
&\Phi(\Gm) := \underset{\wv \in \mathcal{S}_{w}}{\operatorname{\min}}  \ \underset{\uv \in \mathcal{S}_{u}}{\operatorname{\max}} \ \uv^{T} \Gm \wv + \psi( \wv, \uv), \\
&\phi(\gv, \hv) := \underset{\wv \in \mathcal{S}_{w}}{\operatorname{\min}}  \ \underset{\uv \in \mathcal{S}_{u}}{\operatorname{\max}} \ \| \wv \|_2 \gv^{T} \uv - \| \uv \|_2 \hv^{T} \wv + \psi( \wv, \uv), \label{AA2}
\end{align}
\end{subequations}
where $\Gm \in \mathbb{R}^{m \times n}, \gv \in \mathbb{R}^{m}, \hv \in \mathbb{R}^n, \mathcal{S}_w \subset \mathbb{R}^n, \mathcal{S}_u \subset \mathbb{R}^m$ and $\psi : \mathbb{R}^n \times \mathbb{R}^m \mapsto \mathbb{R}$. Denote by $\wv_{\Phi} := \wv_{\Phi}(\Gm) $ and $\wv_{\phi} := \wv_{\phi}( \gv, \hv)$ any optimal minimizers of (\ref{P,AO}) and (\ref{AA2}) respectively. Let $\mathcal{S}_w, \mathcal{S}_u$ be convex, and assume that $\psi(\wv,\uv)$ is convex-concave continuous on $\mathcal{S}_w \times \mathcal{S}_u$ and $\Gm, \gv$ and $\hv $ all have iid standard normal entries. Let $\mathcal{S}$ be any arbitrary open subset of $\mathcal{S}_w $.
Then, if $\lim_{n \rightarrow \infty} \mathbb{P}[\wv_{\phi} \in \mathcal{S}] = 1,$ it also holds that $\lim_{ n \rightarrow \infty} \mathbb{P}[\wv_{\Phi} \in \mathcal{S}] = 1.$

In a nutshell, we study the performance of the (PO) by analyzing its corresponding (AO) which is much easier to study, since it depends on the random vectors $\gv$ and $\hv$ instead of the large random matrix $\Gm$, simplifying the latter to a Scalar Optimization (SO), and finally studying the asymptotic performance of the (SO). To make use of the CGMT, the set $\mathcal{S}$ should be properly chosen as the set in which the MSE concentrates. In the following subsection, we start the MSE analysis by specializing the CGMT to the Box-EN problem at hand and identifying its corresponding (PO) and (AO).
\subsection{Identifying the (PO) and the (AO)}
Under the imperfect measurements assumption, the Box-EN optimization in (\ref{Box-EN_1}) becomes
\begin{equation}\label{Box-EN_2} 
\hat{\xv}=  \text{arg} \ \underset{{\xv \in \mathcal{B}}}{\operatorname{\min}} \ \|  \yv - \Am \xv \|_2^{2} + \lambda_1 \| \xv \|_1+ \lambda_2 \| \xv \|_2^2.
\end{equation}
For convenience, we consider the vector $\wv := \gamma \xv - \xv_0 $, and also the modified Box set:
\begin{equation}
\mathcal{B}' = \{ \wv \in \mathbb{R}^n : l - \xv_{0,i} \leq \wv_i \leq u - \xv_{0,i}, i \in \{1,2, \cdots, n\} \},
\end{equation}
then the problem in (\ref{Box-EN_2}) can be reformulated as
{\footnotesize
\begin{equation*}\label{Lasso_w}
\hat{\wv} = \text{arg} \ \underset{\wv \in \mathcal{B}'}{\operatorname{\min}} \ \| \Hm \wv + \frac{\epsilon}{\gamma} \Omegam(\wv+ \xv_0)-\zv \|_2^2+  \frac{\lambda_1}{\gamma} \| \wv + \xv_0 \|_1+ \frac{\lambda_2}{\gamma^2} \| \wv + \xv_0 \|_2^2.
\end{equation*}}\noindent
This minimization is not in the (PO) form as it is missing the max part. So to go around this, let us express the loss function in its dual form through the Fenchel conjugate, 
$\| \Hm \wv + \frac{\epsilon}{\gamma} \Omegam (\wv+ \xv_0)-\zv \|_2^2$
$= \max_{\uv} \sqrt{n} \uv^T (\Hm \wv + \frac{\epsilon}{\gamma} \Omegam (\wv+ \xv_0)-\zv) - \frac{n}{4} \| \uv \|_2^2$.
Hence, the problem above is equivalent to the following
\begin{align}\label{LASSO-PO1}
&\underset{\wv \in \mathcal{B}'}{\operatorname{\min}} \ \underset{\uv}{\operatorname{\max}} \ \sqrt{n} \uv^T  \Hm \wv +\frac{\sqrt{n} \epsilon}{\gamma} \uv^T \Omegam (\wv+ \xv_0) -\sqrt{n} \uv^T \zv \nonumber \\
& - \frac{n}{4} \| \uv \|_2^2+ \frac{\lambda_1}{\gamma} \| \wv + \xv_0 \|_1 + \frac{\lambda_2}{\gamma^2}\| \wv + \xv_0 \|_2^2.
\end{align}
To reach the desired (PO) form, we introduce the variables
$\vv= \begin{bmatrix}
       \wv            \\[0.3em]
        \frac{\epsilon}{\gamma}(\wv+\xv_0) \\[0.3em]
     \end{bmatrix}$ $\in \mathbb{R}^{2n}$, 
     $\Gm= \begin{bmatrix}
       \Hm &    \Omegam      \\[0.3em]
     \end{bmatrix}$ $\in \mathbb{R}^{m \times 2n}$ and
     $\Cm= \begin{bmatrix}
       \Om_{n} &   \frac{\gamma}{\epsilon} \mathbf{I}_{n}      \\[0.3em]
     \end{bmatrix}$ $\in \mathbb{R}^{n \times 2n}$, where $\Om_{n}$ and $\mathbf{I}_{n}$ are $n \times n$ matrices that represent the all-zeros matrix and the identity matrix, respectively and use them to rewrite (\ref{LASSO-PO1}) in the desired (PO) form as
\small{
\begin{align*}
&\underset{\vv \in \mathcal{D}}{\operatorname{\min}} \ \underset{\uv}{\operatorname{\max}} \ \sqrt{n} \uv^T  \Gm \vv -\sqrt{n} \uv^T \zv - \frac{n}{4} \| \uv \|_2^2 \nonumber\\
& + \frac{\lambda_1}{\gamma} \| \Cm \vv\|_1 + \frac{\lambda_2}{\gamma^2}\| \Cm \vv \|_2^2,
\end{align*}}
\small{
where $\mathcal{D} = \{\vv^T = [ \wv^T \quad \frac{\epsilon}{\gamma}(\wv+\xv_0)^T] : \wv \in \mathcal{B}'\}$. The corresponding (AO) problem is thus given by
\begin{align}\label{L_AO1}
&\underset{\vv \in \mathcal{D}}{\operatorname{\min}} \ \underset{\uv}{\operatorname{\max}} \ \| \vv \|_2 \gv^T \uv - \| \uv \|_2 \muv^T \vv - \frac{n}{4} \| \uv \|_2^2  \nonumber \\
& - \sqrt{n} \uv^T \zv + \frac{\lambda_1}{\gamma} \| \Cm \vv \|_1+ \frac{\lambda_2}{\gamma^2} \| \Cm \vv \|_2^2,
\end{align}\noindent
where $\muv \in \mathbb{R}^{2n}$ and $\gv \in \mathbb{R}^{m}$ are independent standard normal vectors.}
\vspace{-0.5cm}
\subsection{Simplifying the (AO)} 
The next step is to simplify the (AO) to a scalar optimization (SO) problem.
Since the vectors $\gv$ and $\zv$ are independent, 
$  \| \vv \|_2 \gv^T \uv - \sqrt{n} \uv^T \zv$ is equivalent in distribution to $\sqrt{\| \vv \|_2^2+ n \sigma_{\zv}^2}\gv^T \uv$. Therefore, the (AO) can be rewritten as
\small{
\begin{align}\label{simpleAO1}
&\underset{\vv \in \mathcal{D}}{\operatorname{\min}} \ \underset{\uv}{\operatorname{\max}} \ \sqrt{\| \vv \|_2^2+ n \sigma_{\zv}^2}\gv^T \uv - \| \uv \|_2 \muv^T \vv - \frac{n}{4} \| \uv \|_2^2 \nonumber \\
 &  + \frac{\lambda_1}{\gamma} \| \Cm \vv\|_1+ \frac{\lambda_2}{\gamma^2} \| \Cm \vv \|_2^2.
\end{align}}
Let us write (\ref{simpleAO1}) now in terms of the original $\wv$ variable
{\small
\begin{align*}
&\underset{\wv \in \mathcal{B}'}{\operatorname{\min}} \ \underset{\uv}{\operatorname{\max}} \ \sqrt{\| \wv \|_2^2+\frac{\epsilon^2}{\gamma^2} \|\wv + \xv_0 \|_2^2+ n \sigma_{\zv}^2}\gv^T \uv  - \frac{n}{4} \| \uv \|_2^2 \nonumber \\
 & -\|\uv\|_2 (\hv_1^T \wv + \frac{\epsilon}{\gamma} \hv_2^T(\wv +\xv_0)) \nonumber \\
  &+ \frac{\lambda_1}{\gamma} \| \wv+ \xv_0 \|_1+ \frac{\lambda_2}{\gamma^2} \| \wv + \xv_0 \|_2^2,
\end{align*}
\noindent
where $\hv_1, \hv_2 \in \mathbb{R}^{n}$ are independent standard normal vectors.}
\small{Alternatively, since $\wv= \gamma\xv -\xv_0$, we can express the above optimization over $\xv$ instead as
\begin{align*}
&\underset{\xv \in \mathcal{B}}{\operatorname{\min}} \ \underset{\uv}{\operatorname{\max}}  \sqrt{\| \xv \|_2^2+ \| \xv_0\|_2^2- 2 \gamma\xv_0^T \xv  + n \sigma_{\zv}^2}\gv^T \uv - \frac{n}{4} \| \uv \|_2^2\nonumber \\
& - \| \uv \|_2 (\gamma \hv_1 + \epsilon \hv_2)^T \xv + \|\uv\|_2 \hv_1^T \xv_0  + \lambda_1 \| \xv \|_1+ \lambda_2 \| \xv \|_2^2.
\end{align*}}
Fixing the norm of $\uv$ to $\beta: =\| \uv \|_2$, we can easily optimize over its direction by aligning it with $\gv$ further simplifying the (AO) to
\small{
\begin{align}\label{eq:AO_22}
&\underset{\beta \geq 0}{\operatorname{\max}} \ \underset{\xv \in \mathcal{B}}{\operatorname{\min}} \ \sqrt{n} \beta \sqrt{ \frac{1}{n}(\| \xv \|_2^2+ \| \xv_0 \|_2^2-2\gamma \xv_{0}^{T} \xv) + \sigma_{\zv}^2} \| \gv \|_2 \nonumber \\
& - \beta ( \gamma \hv_1 + \epsilon \hv_2)^{T} \xv + \beta \hv_1^{T} \xv_0  - \frac{n \beta^2}{4} + \lambda_1 \| \xv \|_1+  \lambda_2 \| \xv \|_2^2.
\end{align}}
To have a separable optimization problem, we use the following variational form:
$\sqrt{\chi} = \underset{\alpha > 0}{\operatorname{\min}} \ \frac{\alpha}{2} + \frac{\chi}{2 \alpha},$
where 
\begin{equation}\label{chi_eq}
\chi = \frac{1}{n}( \|\xv \|_2^2+ \| \xv_0 \|_2^2- 2 \gamma \xv_{0}^T \xv ) + \sigma_{\zv}^2.
\end{equation}
which reduces (\ref{eq:AO_22}) to
{\small{
\begin{align}\label{AA23}
&\underset{\tau > 0}{\operatorname{\min}} \ \underset{\beta \geq 0}{\operatorname{\max}} \ \frac{ \beta \tau \| \gv \|_2^2}{2} + \frac{n \beta \sigma_{\zv}^2 }{2 \tau} - \frac{n \beta^2}{4} + \frac{\beta \| \xv_0 \|_2^2}{2 \tau}  \nonumber \\
 &+ \beta \hv_1^{T} \xv_0 + \sum_{i = 1}^{n} \biggl( \underset{l \leq \xv_i \leq u}{\operatorname{\min}} \ (\frac{\beta}{2 \tau} + \lambda_2) \xv_i^2 \nonumber \\
&- \beta \biggl( {\hv}_i + \frac{\gamma \xv_{0,i}}{\tau}  \biggr) \xv_i + \lambda_1 |\xv_i| \biggr),
\end{align}}}
where $\tau :=\frac{\sqrt{n} \alpha}{\| \gv \|_2}$, and $\theta= \frac{\beta}{2 \tau} + \lambda_2$. 
We need to prove that the optimal $\beta_*$,
denoted by $\beta_n$, is positive alomst surely.
To do so, suppose by contradiction that $\beta_n$  = 0. First, it is possible to show using
 problem (10) that \small{$\beta_n = 2 \text{max} \{0, \sqrt{\chi}\frac{|| \mathbf g ||}{\sqrt{n}} - \frac{1}{n} \mathbf{ \tilde{h}}^T \mathbf x + \frac{1}{n} \mathbf h_1^T \mathbf  x_0\}$}, where $\mathbf {\tilde{h}} = \gamma \mathbf h_1 + \epsilon \mathbf h_2$. If $\beta_n=0$, then this is equivalent to say that

\begin{equation}
\sqrt{\chi}\frac{|| \mathbf g ||}{\sqrt{n}} - \frac{1}{n} \mathbf{ \tilde{h}}^T \mathbf x + \frac{1}{n} \mathbf h_1^T \mathbf  x_0\ \leq 0.
\end{equation}
If $\beta_n=0$, the problem in (12) simplifies to the following: $\sum_{i=1}^{n} \underset{l \leq \mathbf x_i \leq u}{\text{min}} \lambda_2 \mathbf x_i^2 + \lambda_1 | \mathbf x_i |$, which implies that the optimal $\mathbf x$ is the zero vector. Consequently, $\sqrt{\chi}\frac{|| \mathbf g ||}{\sqrt{n}} - \frac{1}{n} \mathbf{ \tilde{h}}^T \mathbf x +\frac{1}{n} \mathbf h_1^T \mathbf  x_0 = \sqrt{\frac{1}{n} || \mathbf {x}_0 ||^2 + \sigma_{\mathbf z}^2} \frac{|| \mathbf {g} ||}{\sqrt{n}} + \frac{1}{n}\mathbf {h}_1^T \mathbf x_0$. Recall that $\mathbf {h}_1$ is a standard Gaussian
vector and then $\frac{1}{n} \mathbf {h}_1^T \mathbf {x}_0$ converges almost surely to zero. Hence, the latter quantity is almost surely strictly
positive, which contradicts (13), i.e. the assumption $\beta_n = 0$.
Thus, $\beta_n  > 0$ almost surely.
After some algebraic manipulations, we can write
\begin{align*}
&\underset{\tau > 0}{\operatorname{\min}} \ \underset{\beta > 0}{\operatorname{\max}} \ \frac{ \beta \tau \| \gv \|_2^2}{2} + \frac{n \beta \sigma_{\zv}^2 }{2 \tau} - \frac{n \beta^2}{4} 
 + \beta \hv_1^{T} \xv_0  \nonumber \\
 &+  \sum_{i = 1}^{n} (\frac{\beta}{2 \tau} - \frac{\beta^2 \gamma^2}{4 \theta \tau^2}) \xv_{0,i}^2 - \frac{\beta^2 \gamma}{2 \theta \tau} {\hv}_i \xv_{0,i} - \frac{\beta^2}{4 \theta} {\hv}_{i}^2  \nonumber \\
&+ 2 \theta \biggr( \sum_{i = 1}^{n} \underset{l \leq \xv_i \leq u}{\operatorname{\min}} \ \frac{1}{2} ( \xv_i -\frac{\beta}{2 \theta} (\frac{\gamma }{\tau}\xv_{0,i} + {\hv}_i) )^2 + \frac{\lambda_1}{2 \theta} |\xv_i| \biggl).
\end{align*}The optimization over $\xv_i$ can be solved in closed-form using the \textit{saturated} soft-thresholding operator $\eta(a ; \lambda, l, u) =\text{argmin}_{l \leq x \leq u}$ $\frac{1}{2}(x-a)^2 + \lambda |x|$ defined as: 
\small{
\begin{equation}\label{soft_TH}
\eta(a ; \lambda, l, u) =
\begin{cases} 
         u & ,\text{if}  \ a > u + \lambda \\ 
           
         a - \lambda & ,\text{if}  \ \lambda < a < u + \lambda \\      
   
		 0 & ,\text{if} \  |a| \leq \lambda  \\
			
		 a + \lambda & ,\text{if}  \ l - \lambda < a < - \lambda \\      

         l  & ,\text{if} \  a \leq l - \lambda.
\end{cases}
\end{equation}
Also, let $e(a;\lambda,l,u) =\text{min}_{l \leq x \leq u} \frac{1}{2}(x-a)^2 + \lambda |x|$ which is defined as:
\begin{equation}
e(a ; \lambda, l, u) =
\begin{cases} 
      \frac{1}{2} (u - a)^2 + \lambda u  & ,\text{if}  \ a > u + \lambda \\      
					
	   \lambda a - \frac{1}{2} \lambda^2  & ,\text{if}  \ \lambda < a < u + \lambda \\     
	    
			\frac{1}{2} a^2 & ,\text{if} \  |a| \leq \lambda   \\
			
      -\lambda a - \frac{1}{2} \lambda^2  & ,\text{if} \  l - \lambda < a < -\lambda \\
      
            \frac{1}{2} (l - a)^2 - \lambda l & ,\text{if} \  a \leq l - \lambda. 
\end{cases}
\end{equation}
}
Then, the above optimization problem finally simplifies to the following Scalar Optimization (SO) problem:
\begin{align}\label{AO22}
&\underset{\tau > 0}{\operatorname{\min}} \ \underset{\beta > 0}{\operatorname{\max}} \ \tilde{D}(\tau,\beta,\gv,\hv) := \frac{ \beta \tau \| \gv \|_2^2}{2} + \frac{n \beta \sigma_{\zv}^2 }{2 \tau} - \frac{n \beta^2}{4} 
  \nonumber \\
 &+  \sum_{i = 1}^{n} (\frac{\beta}{2 \tau} - \frac{\beta^2 \gamma^2}{4 \theta \tau^2}) \xv_{0,i}^2 +(\beta- \frac{\beta^2 \gamma}{2 \theta \tau}) \hv_i \xv_{0,i} - \frac{\beta^2}{4 \theta} \hv_{i}^2  \nonumber \\
&+ 2 \theta  \sum_{i = 1}^{n} e \biggr( \frac{\gamma \beta}{2 \theta \tau}\xv_{0,i} + \frac{\beta}{2 \theta} \hv_i ; \frac{ \lambda_1}{2 \theta}, l ,u \biggl).
\end{align}
\subsection{Probabilistic asymptotic analysis of the (SO) problem}
After simplifying the (AO) to its form in (\ref{AO22}), we are now in a position to analyze its limiting behavior.
First, we need to properly normalize the objective function in (\ref{AO22}) by dividing it by $n$. 
Then, using the weak law of large numbers (WLLN), we have\footnote{We write ``$\overset{P}{\longrightarrow} $" to designate convergence in probability.}: $\frac{1}{n} \| \gv \|_2^2\overset{P}{\longrightarrow} \delta, \frac{1}{n} \| \hv \|_2^2\overset{P}{\longrightarrow} 1$, $\frac{1}{n} \| \xv_0\|_2^2\overset{P}{\longrightarrow} \kappa $ and $\frac{1}{n} \hv^T \xv_0 \overset{P}{\longrightarrow} 0$. Also, using the WLLN, it can be shown that for all $\tau>0$ and $\beta > 0$,
$\frac{1}{n} \sum_{i = 1}^{n} e ( \frac{\gamma \beta}{2 \theta \tau}\xv_{0,i} + \frac{\beta}{2 \theta} \hv_i ; \frac{ \lambda_1}{2 \theta}, l ,u ) \overset{P}{\longrightarrow} \mathbb{E} [e (\frac{\gamma \beta}{2 \theta \tau} X_0 + \frac{\beta}{2 \theta} H ; \frac{ \lambda_1 }{2 \theta},l,u ) ]$, and $\frac{1}{n} \sum_{i=1}^{n} \tilde{\xv}_i \overset{P}{\longrightarrow} \mathbb{E} [\eta (\frac{\gamma \beta}{2 \theta \tau} X_0 + \frac{\beta}{2 \theta} H ; \frac{ \lambda_1 }{2 \theta},l,u ) ]$, where $\tilde{\xv}$ is the solution of (AO) defined in (\ref{eq:AO_22}) and $H$ is a scalar standard normal random variable.
Therefore, the point-wise convergence in $\tau$ and $\beta$ of the objective function in (\ref{AO22}) is 
$D(\tau, \beta):= \frac{ \beta \tau  \delta }{2} + \frac{ \beta \sigma_{\zv}^2 }{2 \tau} - \frac{ \beta^2}{4\theta}(\theta+1)+ (\frac{ \beta }{2 \tau}  - \frac{\beta^2 \gamma^2}{4 \theta \tau^2}) \kappa   + 2 \theta  \mathbb{E}_{X_0,H}[e (\frac{\gamma \beta}{2 \theta\tau} X_0 + \frac{\beta}{2 \theta} H ; \frac{ \lambda_1 }{2 \theta},l,u ) ]$. Furthermore, it is possible to show that with probability one, the functions 
$\tau \mapsto {\operatorname{\max}}_{\beta>0} \tilde{D}(\tau,\beta,\gv, \hv)$ and $\tau \mapsto {\operatorname{\max}}_{\beta>0} D(\tau,\beta)$ are convex in $\tau$. Hence, it is possible to show using Theorem 2.7 in \cite{newey1994large} that $\tau_n(\gv,\hv) \overset{P}{\longrightarrow} \tau_*$. Likewise, it can be similarly proved that $\beta_n(\gv,\hv) \overset{P}{\longrightarrow} \beta_*$.
\subsection{Applying the CGMT}
Now, we are in a position to study the convergence limit of the MSE of the Box-EN. First, recall from (\ref{chi_eq}) that $\frac{1}{n} \| \tilde{\xv} - \xv_0 \|_2^2= \frac{\| \gv \|_2^2}{n} \tau_n^2(\gv,\hv) + \frac{2}{n}(\gamma-1) \tilde{\xv}^T \xv_0 -\sigma_{\zv}^2$. Moreover, it can be shown that $\frac{1}{n}\tilde{\xv}^T \xv_0 \overset{P}{\longrightarrow} \mathbb{E}[ \eta (\frac{\gamma \beta_*}{2 \theta_* \tau_*} X_0 + \frac{\beta_*}{2 \theta_{*}} H ; \frac{ \lambda_1 }{2 \theta_*},l,u)  X_0]:= \Upsilon(\tau_*,\beta_*)$. Using $\tau_n(\gv,\hv)  \overset{P}{\longrightarrow} \tau_*$, and $\beta_n(\gv,\hv)  \overset{P}{\longrightarrow} \beta_*$, we can show that: 
\begin{align}\label{M-eq}
\frac{1}{n} \| \tilde{\xv} - \xv_0 \|_2^2&\overset{P}{\longrightarrow}  \delta \tau_*^2 - \sigma_{\zv}^2 + 2(\gamma -1)\Upsilon(\tau_*,\beta_*)   \nonumber \\
&:= M(\tau_*, \beta_*).
\end{align}
The last step is to use the CGMT to prove that the quantities $\hat{\xv} - \xv_0$ and $\tilde{\xv} - \xv_0$ are concentrated in the same set. Formally,
for any fixed $\zeta > 0$, we define the set
$\mathcal{S} = \bigl \{ \rv: \bigl|  \frac{1}{n}\| \rv \|_2^2- M(\tau_*,\beta_*)  \bigr| < \zeta \bigr\}$.\\
Equation (\ref{M-eq}) proves that for any $\zeta>0$, $\tilde{\xv} - \xv_0 \in \mathcal{S}$ with probability one. 
Then, we conclude using the CGMT that $\hat{\xv} - \xv_0 \in \mathcal{S}$ with probability one.
The asymptotic prediction of the MSE is summarized in Theorem \ref{EN_mse} given in the next section. 
\section{Main Results}
\label{sec:Results}
This section summarizes our main results on the precise analysis of the mean squared error and the probability of support recovery of the Box-EN.
We use standard notation $\textsl{plim} \ X_n = X$ to denote that a sequence of random variables
$X_n$ converges in probability towards a constant $X$.
\vspace{-0.2cm}
\begin{theorem}[Box-EN MSE]\label{EN_mse}
Fix $\lambda_1, \lambda_2 > 0$, and let $\hat{\xv}$ be a minimizer of the Box-EN problem in (\ref{Box-EN_2}), where $\Am, \zv$ and $\xv_0$ satisfy the working assumptions of Section \ref{w_A}.
Then, in the limit of
$m, n \to +\infty, m/n = \delta$, it holds:
\small{
\begin{align}
&\textsl{plim} \ \frac{1}{n} \| \hat{\xv} - \xv_0 \|_2^2= \delta \tau_{*}^2 - \sigma_{\zv}^2 \nonumber \\
&+2(\gamma -1)\mathbb{E}_{\underset{H \sim \mathcal{N}(0,1)}{X_0 \sim p_{X_0}} } \biggr[\eta \biggr(\frac{\gamma \beta_*}{2 \theta_* \tau_*} X_0 + \frac{\beta_*}{2 \theta_{*}} H ; \frac{ \lambda_1 }{2 \theta_*},l,u \biggl)  X_0  \biggl] ,
\end{align}}
where 
$\theta_* = \frac{\beta_*}{2 \tau_*} +  \lambda_2$, and $(\tau_*,\beta_*)$ is the unique solution to the following:
\small{
\begin{align} \label{optimal_t_b}
&\underset{\tau > 0}{\operatorname{\min}} \ \underset{\beta > 0}{\operatorname{\max}} \ D(\tau, \beta ):= \frac{ \beta \tau  \delta }{2} + \frac{ \beta \sigma_{\zv}^2 }{2 \tau} - \frac{ \beta^2}{4\theta}(\theta+1) \nonumber \\
&+ (\frac{ \beta }{2 \tau}  - \frac{\beta^2 \gamma^2}{4 \theta \tau^2}) \kappa 
  + 2 \theta \mathbb{E}_{X_0,H} \biggr[e \biggr(\frac{\gamma \beta}{2 \theta\tau} X_0 + \frac{\beta}{2 \theta} H ; \frac{ \lambda_1 }{2 \theta},l,u \biggl) \biggl].
\end{align}}
\end{theorem}
\noindent
\textit{Remark} 1 (Optimal Solutions). Note that $\tau_*$ and $\beta_*$ can be efficiently computed by writing the first order stationary point conditions, i.e., $\nabla_{(\tau,\beta)}  D(\tau, \beta )=0.$\\
\noindent
\textit{Remark} 2 (Optimal Regularizers). Theorem 1 can be used to find the optimal pair of regularizers $(\lambda_1,\lambda_2)$ that minimizes the MSE.
\noindent
\textit{Remark} 3. For a sparse Bernoulli vector $\xv_0$, the expectation in (18) is given by:
\begin{align}
&\mathbb{E} \biggr[\eta \biggr(\frac{\gamma \beta_*}{2 \theta_* \tau_*} X_0 + \frac{\beta_*}{2 \theta_{*}} H ; \frac{ \lambda_1 }{2 \theta_*},l,u \biggl)  X_0  \biggl] = \kappa u Q(\frac{2 \theta_* (u-\xi) + \lambda_1 }{\beta_*}) \nonumber\\
&+ \kappa l Q(\frac{2 \theta_* (\xi-l) + \lambda_1 }{\beta_*}) +\kappa \int_{\frac{\lambda_1 -2\theta_* \xi}{\beta_*}}^{\frac{2 \theta_*(u-\xi)+\lambda_1}{\beta_*}} (\xi + \frac{\beta_* h -\lambda_1}{2 \theta_*}) p(h) dh \nonumber\\
&+\kappa \int_{\frac{ 2\theta_* (l-\xi)-\lambda_1}{\beta_*}}^{\frac{-2 \theta_*\xi-\lambda_1}{\beta_*}} (\xi + \frac{\beta_* h +\lambda_1}{2 \theta_*}) p(h) dh,
\end{align}
where $p(x) = \frac{1}{\sqrt{2 \pi}} e^{-x^2/2}$ is the pdf of a standard Gaussain random variable and $Q(x)$ is its associated Q-function.
The expectation in (19) can be found in a similar way.

The following Theorem precisely characterizes the support recovery metrics introduced in (\ref{supp}).
\begin{theorem}[Probability of support recovery]\label{EN_on/off}
Under the same settings of Theorem \ref{EN_mse} and for any fixed $\xi>0$,
and in the limit of
$m, n \to +\infty, m/n = \delta$, it holds that:
\small{
\begin{equation}\label{on_TH}
\textsl{plim} \ \Phi_{\xi,on}(\hat{\xv}) = \mathbb{P} \biggl[\biggl | \eta \biggr(\frac{\gamma \beta_*}{2 \theta_* \tau_*} X_0 + \frac{\beta_*}{2 \theta_*} H ; \frac{ \lambda_1 }{2 \theta_*},l,u \biggl)  \biggr |   \geq \xi \biggr], 
\end{equation}
and 
\begin{equation}\label{off_TH}
\textsl{plim} \ \Phi_{\xi,\text{off}}(\hat{\xv}) = \mathbb{P} \biggl[ \biggl| \eta \biggr( \frac{ \beta_*}{2 \theta_*} H ; \frac{ \lambda_1 }{2 \theta_*},l,u \biggl)  \biggr|   \leq \xi \biggr].
\end{equation}
}
\end{theorem}
\noindent
The proof of Theorem \ref{EN_on/off} is also based on the CGMT and largely follows the proof of Theorem \ref{EN_mse} but is omitted for space limitations (see \cite{thrampoulidis2018symbol} for a similarly detailed treatment).\\
\noindent
\textit{Remark} 1 (Small/Large Regularizers). From Theorem 2, it can be seen that as $\lambda_1$ becomes large $\Phi_{\xi,\text{off}}(\hat{\xv})$ converges to one while $\Phi_{\xi,\text{on}}(\hat{\xv})$ converges to zero. Opposite behavior is observed when $\lambda_1$ takes values close to zero.
This behavior is expected since large values of $\lambda_1$ parameter put more emphasis on the $\ell_1$-norm term, thus promoting sparser solution. This is clearly depicted in Figure 3.\\
\noindent
\textit{Remark} 2 (Optimal Regularizers).
In order to trade-off between on- and off- support recovery
probabilities a reasonable performance metric will be $\Phi_{\xi} =\omega \Phi_{\xi,\text{on}}+ (1-\omega) \Phi_{\xi,\text{off}}$ for $\omega \in [0, 1]$. Theorem 2 precisely characterizes the behavior of this as a function of $(\lambda_1,\lambda_2)$; thus, it determines the optimal value of $(\lambda_1,\lambda_2)$ that minimizes $\Phi_{\xi}$ \cite{abbasi2016general}.\\
\noindent
\textit{Remark} 3.
The parameters $l$ and $u$ generally do not need to be tuned because the support of the signal is known (take the example of digital communication systems where the constellation is known by both the transmitter and the receiver). Hence, the choice of $l$ and $u$ is straightforward, i.e. $l $ = min $ \mathbf x_0$ and $u$ = max $ \mathbf x_0$.\\
\noindent
\textit{Remark} 4. For a sparse Bernoulli vector $\xv_0$, the expressions in (\ref{on_TH}), and (\ref{off_TH}) simplify to  
\begin{align}
\textsl{plim} \ \Phi_{\xi,\text{on}}(\hat{\xv}) = \ \left(\frac{2 \theta_* \xi + \lambda_1}{\beta_*} - \frac{\gamma }{\tau_*} \right),
\end{align}
and 
\begin{equation}
\textsl{plim} \ \Phi_{\xi,\text{off}}(\hat{\xv}) = 1 - Q\left(\frac{2 \theta_* \xi + \lambda_1}{ \beta_*} \right).
\end{equation}
\\
\noindent
\textit{Remark} 5. Theorems 1 and 2 can also be used to characterize the performance of the standard EN  by letting $l \to -\infty$, and $u \to \infty$, in which the saturated soft-thresholding operator reduces to the conventional soft-thresholding operator.
To the best of our knowledge, the precise analysis of the standard EN under matrix uncertainties has not been studied before as well. However, we focus in this paper on the Box-EN since it generalizes the standard EN and gives better performance as compare to the standard one. 
\vspace{-0.3cm}
\section{Numerical Results}
For illustration, we focus only on the case where $\xv_0$ has entries that are sampled iid from a sparse-Bernoulli distribution with $\mathbb{P}[X_0 = 0] =0.9$ and $\mathbb{P}[X_0 = 1] =0.1$ (i.e. $\kappa=0.1$). A natural choice for the box-constraint values in this situation is to set $l=0$, and $u=1$.
Figure \ref{mse_Fig} shows the close agreement between the asymptotic mean squared error (MSE) of the Box-EN as predicted by Theorem \ref{EN_mse} and numerical simulations. The simulation results are averages over 100 realizations
of $\Am$ and $\zv$ with $n=500$ and $\delta = 0.7$. This figure also demonstrates that the Box-EN outperforms the standard EN. \\
From Figure \ref{mse_Fig}, we can see that as the regularizer $\lambda_2$ is varied, we notice a pronounced minimum for some $\lambda_2> 0$.
\begin{figure}
\begin{center}
\includegraphics[width=8cm, height =6.2cm]{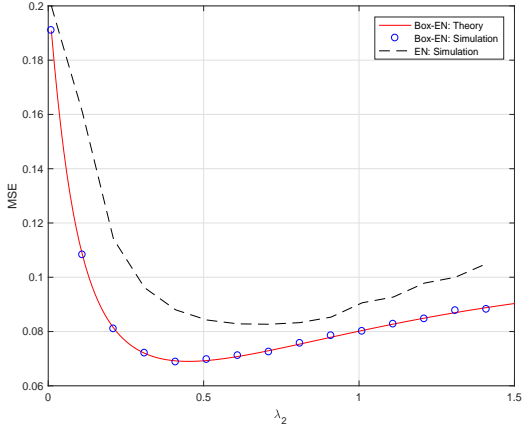}
\caption{\scriptsize{MSE of the Box-EN and the EN. This figure shows that the former outperforms the latter. Theoretical prediction is based on Theorem \ref{EN_mse} with $\lambda_1 =0.1$. For the simulations, we used $\kappa =0.1, \epsilon^2= 0.1,\delta = 0.7,n=500$, SNR = 0.5.}}%
\label{mse_Fig}
\end{center}
\end{figure}
We also we plotted the MSE as a function of both $\lambda_1$, and $\lambda_2$ in Figure \ref{mse_3D}. The marked point is the point corresponding to the minimum value of MSE.
\begin{figure}
\begin{center}
\includegraphics[width=8cm, height =6.2cm]{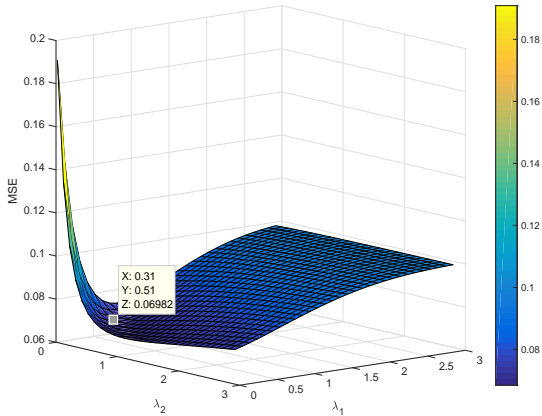}
\caption{\scriptsize{MSE of the Box-EN as a function of $\lambda_1, \lambda_2$ for a sparse Bernoulli signal. We used $\delta =0.7, \kappa = 0.1,\epsilon^2 = 0.1,n=500$, and SNR = 0.5.}}%
\label{mse_3D}
\end{center}
\end{figure}
As we can see from this figure, the optimal pair $(\lambda_1, \lambda_2)$ that minimizes the MSE is non-zero for both $\lambda_1$, and $\lambda_2$.
\begin{itemize}
\item  \textbf{For $\epsilon^2$}, which is the variance of the estimation error, as $\epsilon^2$ increases we get worse and worse MSE values as shown in Figure 4 below. This is clear, since poor estimation of the measurement matrix would result in a poor signal recovery.
\begin{figure}
\begin{center}
\includegraphics[width=8cm, height =6.2cm]{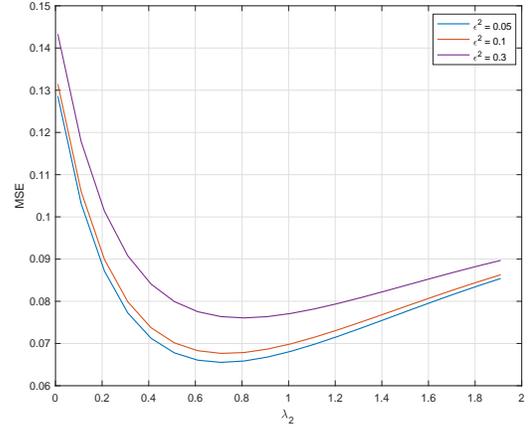}
\caption{\scriptsize{MSE of the Box-EN for different values of $\epsilon^2$. We used $\lambda_1 =0.1, \kappa =0.1,\delta = 0.8,n=500$, SNR = 0.5.}}%
\label{mse_Fig}
\end{center}
\end{figure}
\item \textbf{For $\delta$}, which is defined as $\delta = \frac{m}{n}$, we plotted (in Figure 5) the MSE for a different values of $\delta$ while fixing all other parameters. The MSE is better as $\delta$ increases. For $\delta <1$, we have an underdetermined system which makes the signal recovery more challanging than the case of an overdetermined system which corresponds to $\delta >1$.
\begin{figure}
\begin{center}
\includegraphics[width=8cm, height =6.2cm]{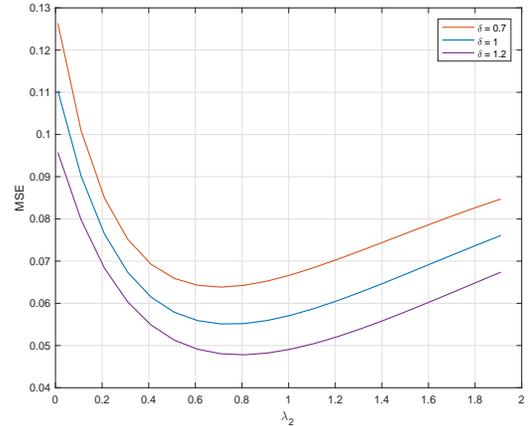}
\caption{\scriptsize{MSE of the Box-EN for different values of $\delta$. We used $\lambda_1 =0.1, \kappa =0.1,\epsilon^2 = 0.1,n=500$, and SNR = 0.5.}}%
\end{center}
\end{figure}
\begin{figure}[h!]%
\begin{center}
\includegraphics[width=8cm, height =6.1cm]{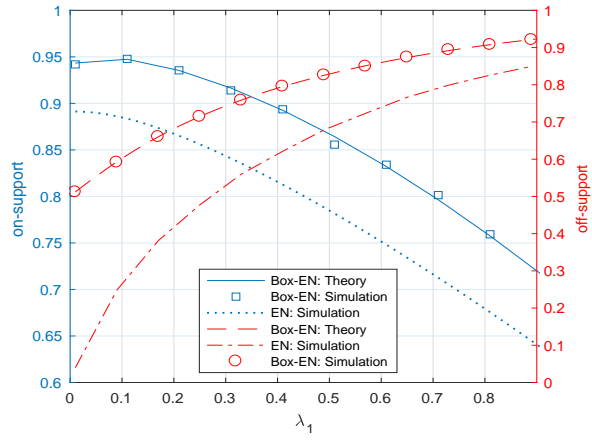}
\caption{\scriptsize{Probability of successful support recovery of the Box-EN and the EN. This figure again shows that the former outperforms the latter.  Theoretical prediction is based on Theorem \ref{EN_on/off} with $\lambda_2 =0.1$, and $\xi =10^{-3}$. For the simulations, we used $\kappa =0.1, \epsilon^2= 0.1,\delta = 0.7,n=500$, SNR = 0.5.}}%
\label{fig:on/off}
\end{center}
\end{figure}
\end{itemize}
The prediction of Theorem \ref{EN_on/off} for the probability of support recovery is compared with the corresponding values obtained via numerical simulations and is plotted in Figure \ref{fig:on/off}. This figure again shows the high accuracy of the proposed theoretical predictions.
\section{Conclusions}
\label{sec:conc}
In this paper, we have used the recently developed Convex Gaussian Min-max Theorem (CGMT) framework to derive precise asymptotic characterizations of the mean squared error (MSE) and probability of support recovery of the Box-Elastic Net (Box-EN) under the presence of uncertainties in the measurement matrix. Numerical simulations show the close agreement to the proposed theoretical predictions. Also, we showed that the Box-EN outperforms standard Elastic-Net. These predictions can be used to optimally tune the involved parameters of the algorithm. Finally, we note that following the same steps as in this paper, our results can be used to characterize the performance of the standard EN.
\newpage
\clearpage
\bibliographystyle{IEEEbib}
\bibliography{References}
\end{document}

%% file: macros.tex
\long\def\comment#1{}

\newfont{\bbb}{msbm10 scaled 700}

\newfont{\bb}{msbm10 scaled 1100}

\newcommand{\gv}{{\bf g}}
\newcommand{\hv}{{\bf h}}

\newcommand{\rv}{{\bf r}}

\newcommand{\uv}{{\bf u}}
\newcommand{\wv}{{\bf w}}
\newcommand{\vv}{{\bf v}}
\newcommand{\xv}{{\bf x}}
\newcommand{\yv}{{\bf y}}
\newcommand{\zv}{{\bf z}}

\newcommand{\Am}{{\bf A}}

\newcommand{\Cm}{{\bf C}}

\newcommand{\Gm}{{\bf G}}
\newcommand{\Hm}{{\bf H}}

\newcommand{\Om}{{\bf O}}

\newcommand{\muv}{\hbox{\boldmath$\mu$}}

\newcommand{\Omegam}{\hbox{\boldmath$\Omega$}}